# Event-based dataset for the detection and classification of manufacturing assembly tasks

Laura Duarte*,[1], Pedro Neto[1]

1 - Centre for Mechanical Engineering, Materials and Processes (CEMMPRE), ARISE, University of Coimbra, 3030-788, Coimbra, Portugal

**Abstract**

The featured dataset, the **Event-based Dataset of Assembly Tasks (EDAT24)**, showcases a selection of manufacturing primitive tasks (idle, pick, place, and screw), which are basic actions performed by human operators in any manufacturing assembly. The data were captured using a DAVIS240C event camera, an asynchronous vision sensor that registers events when changes in light intensity value occur. Events are a lightweight data format for conveying visual information and are well-suited for real-time detection and analysis of human motion. Each manufacturing primitive has 100 recorded samples of DAVIS240C data, including events and greyscale frames, for a total of 400 samples. In the dataset, the user interacts with objects from the open-source CT-Benchmark in front of the static DAVIS event camera. All data are made available in raw form (.*aedat*) and in pre-processed form (.*npy*). Custom-built Python code is made available together with the dataset to aid researchers to add new manufacturing primitives or extend the dataset with more samples.

## SPECIFICATIONS TABLE

| Subject | Computer Vision and Pattern Recognition |
|---|---|
| **Specific subject area** | Human action/task classification in a manufacturing assembly scenario. |
| **Data format** | Raw event camera data (.*aedat*) <br><br> Text file (.*csv*) <br><br> Data structure file (.*npy*) |
| **Type of data** | DAVIS240C event camera data (.*aedat*), Raw timestamp data (.*csv*), <br><br> Event data in (x,y,ts,pol) format (.*npy*) |
| **Data collection** | Data were collected with a DAVIS240C event camera in a static position. Custom build Python code is used in conjunction with the jAER software to carry out the data collection process. |
| **Data source location** | Institution: Department of Mechanical Engineering, University of Coimbra <br><br> City: Coimbra |

|  | Country: Portugal |
|---|---|
| **Data accessibility** | Data available at Zenodo |
|  | Repository name: **Event-based Dataset of Assembly Tasks (EDAT24)** |
|  | Data identification number (DOI): 10.5281/zenodo.10562563 |
|  | Direct URL to data: https://zenodo.org/records/10562563 |
|  | Code available at GitHub |
|  | Repository name: **DAVIS-data-capture-system** |
|  | Data identification number (DOI): 10.5281/zenodo.10569638 |
|  | Direct URL to code: https://github.com/Robotics-and-AI/DAVIS-data-capture-system |

# VALUE OF THE DATA

- Due to its asynchronous nature, event data have many advantages such as high temporal resolution and low latency [1]. However, datasets using event data are still scarce, especially when applied to the manufacturing domain.
- Event data are made available in raw form (*.aedat*) and in pre-processed form (*.npy*), so researchers can choose how to apply their event-based classification methods.
- The data are collected in a replicable setup with an open-source benchmark developed specifically for human-robot collaboration scenarios.
- The dataset's data and structure lend themselves to the addition of more data.
- Event data can be used to train and accurately classify human actions [2]. This is essential to enable robots to anticipate/predict human actions, recognize the need for help, and guarantee a safe environment for human-robot collaboration.
- The dataset is suitable for anyone conducting research activities in computer vision and machine learning using event cameras aiming to create and test event-based algorithms. The focus lies on tracking and classifying human action primitives, an ongoing research topic [3].

# BACKGROUND

Machine vision sensing is often employed for capturing environment data within robotic systems. Traditional frame-based cameras can produce high-quality images suitable for feeding powerful machine learning algorithms, which can be used, for example, to classify human actions and objects. Nonetheless, their performance is lacking in high-speed applications as they suffer from motion-blur, relative high latencies, and limited dynamic range. Event cameras were developed as a new vision sensor to face the frame-based cameras' challenges. An event camera works asynchronously, resulting in high temporal resolution and low latency, which is crucial for detection and analysis of motion. Event cameras also have a high dynamic range because events are registered when changes in logarithmic light intensity *log(I)* occur. Due to its novelty, event-based classification methods have room for improvement, requiring new data in the form of datasets. In the field of human action classification, there are currently only a small set of event-based datasets. This encourages the creation of novel event-based datasets focused on manufacturing tasks.

## DATA DESCRIPTION

The EDAT24 dataset contains a total of 400 sample videos in *.aedat* (AEDAT 2.0) format, Figure 1. The AEDAT 2.0 (Address Event DATa) file format stores both frame data and event information. Events are stored in series of [address, timestamp] pairs, each 32 bits wide, for a total of 8 bytes per event. The address must be interpreted to obtain relevant event information (the event's x coordinate, y coordinate, and polarity) and the timestamp is registered in microseconds. The events acquisition rate of the dataset videos is, on average, 1.65M (million) Hz.

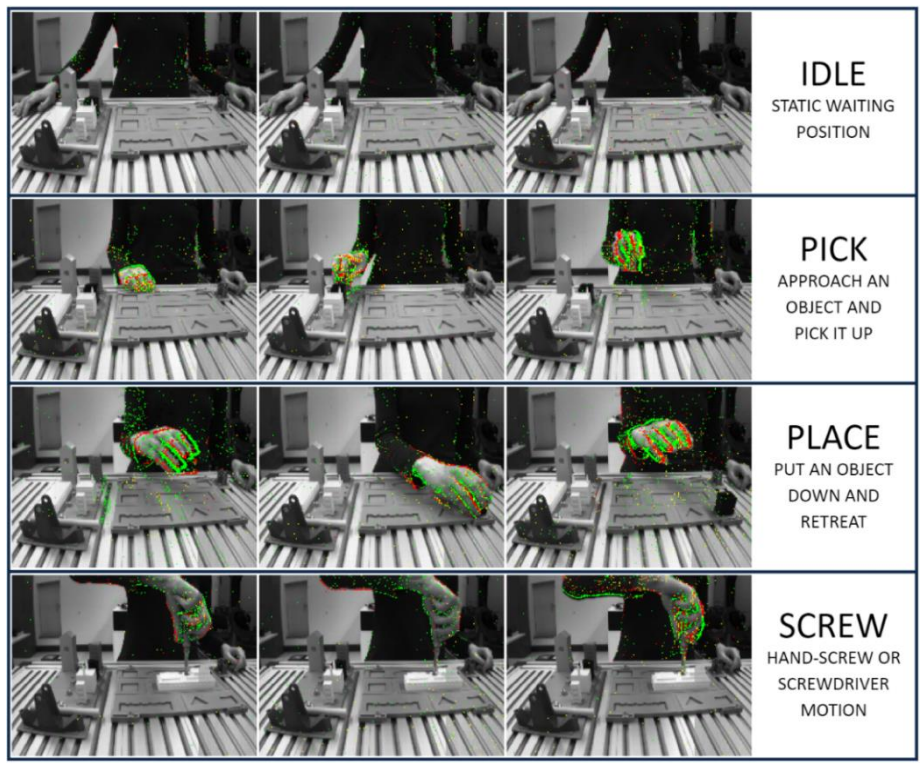

Figure 1: Showcase of frames and events of the EDAT24 dataset, captured by the DAVIS240C. The positive and negative polarity events are overlayed over each frame as green and red points, respectively.

These raw videos are labelled using the format ***primitive/task_nSample.aedat***, where:

- ***primitive*** corresponds to the various primitive tasks (**idle, pick, place, and screw**).
- ***task*** corresponds to the name of the part of the CT-Benchmark the user is interacting with.
- ***nSample*** corresponds to the sample number for each task.

For example, the file named ***pick/bridge_peg_2.aedat*** corresponds to the video sequence of the second performance of the pick movement of a bridge peg part, Figure 2.

For each *.aedat* file, a corresponding *.csv* (comma-separated values) file of the same name contains the timestamps with the start and end time of the recording in microseconds. The timestamp of the first and last event are recorded when the trigger is pressed to start and stop the recording, containing precise information about the duration of each recording.

The NumPy file format inherently has fast loading times and small file sizes due to being a standard binary file format. As such, to improve the dataset usability, especially for those who want to use already processed event data, all *.aedat* files are individually processed to obtain event-only data and made available in *.npy* (NumPy) files. This data includes a list of all event x and y coordinates, a list of event polarities and a list of event timestamps.

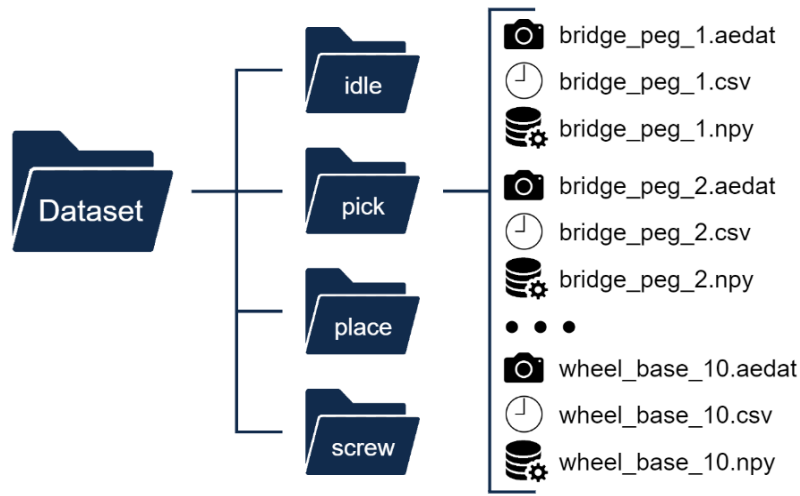

Figure 2: EDAT24 dataset directory structure.

# EXPERIMENTAL DESIGN, MATERIALS AND METHODS

The data were collected from manufacturing assembly tasks within the Collaborative Task Benchmark (CT-Benchmark) [4], Figure 3. The assembly of the building models allows for different assembly options using various tasks, i.e., the manufacturing primitives in EDAT24 (idle, pick, place, and screw). Only some of the benchmark building models and related tasks were used to create the EDAT24 dataset, namely the tasks assigned to the human in a human-robot collaborative assembly[1].

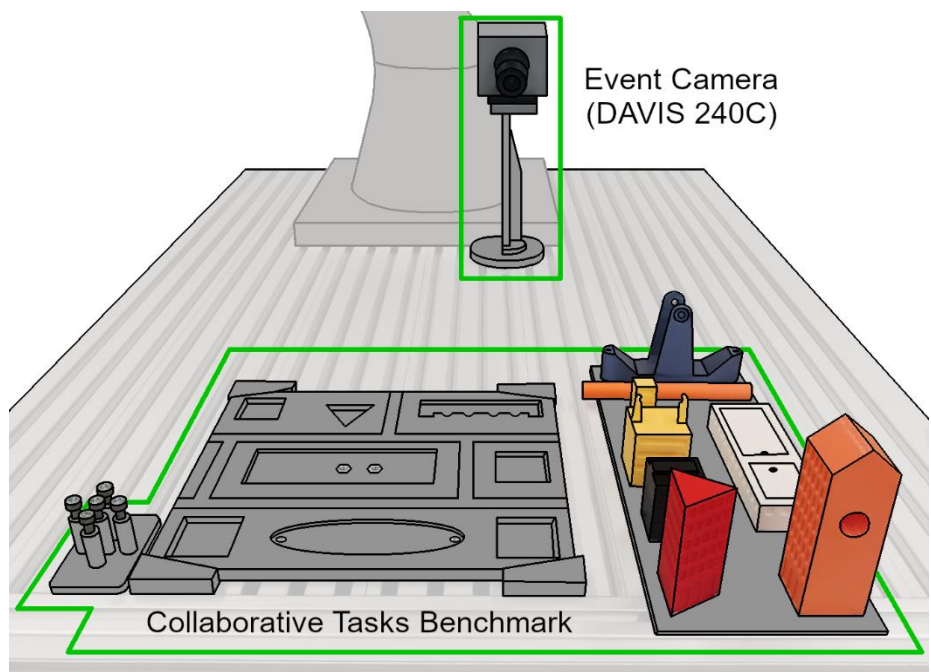

Figure 3: Setup of the workbench for obtaining data for the EDAT24 dataset.

---

[1] https://github.com/Robotics-and-AI/collaborative-tasks-benchmark/tree/main/videos

After preparing the setup of the CT-Benchmark, the DAVIS240C event camera [5] is placed at a fixed point in the workspace to capture the dataset videos, Figure 4. It is connected to a computer through a USB cable and the open-source software jAER is used to capture and visualize data [6]. Since the event camera is in a fixed position, stationary elements will not be registered in the event data. As such, the event camera only captures event data from the human hands and torso moving relative to the camera, Figure 1. Additionally, the DAVIS240C was set to simultaneously capture grayscale frames at a rate of 20 frames per second. These grayscale frames are also shared in the dataset and accessible through decoding the *.aedat* file.

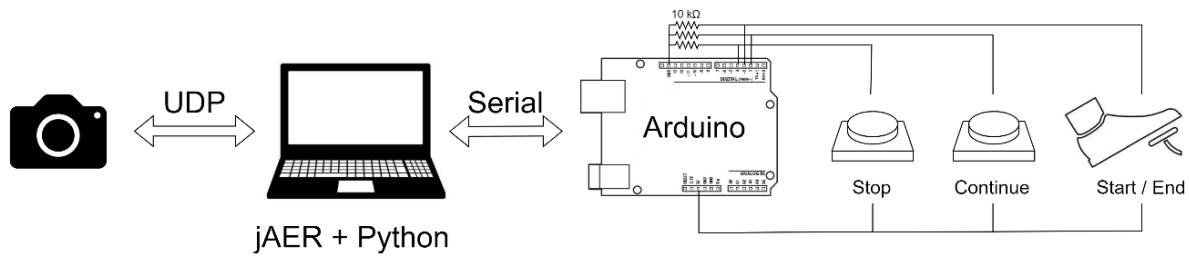

Figure 4: The dataset acquisition hardware and software setup.

The hardware setup comprises an Arduino connected to a foot pedal and two push buttons to signal the capture system to start and end the recordings. The Arduino is linked to the computer through a USB cable. Upon pressing the foot pedal, a signal is transmitted to the Arduino, allowing the user to effortlessly start and stop recordings. Two supplementary buttons enable the user to choose between continuing or interrupting the recording process.

Custom-built Python software handles communication with the event camera (through jAER), communication with Arduino, data recording, and creation of all files in the dataset, Figure 4. This software is freely available in a GitHub Repository [7]. The repository also contains the Arduino code[2] and a video demonstrating the use of this software for data acquisition[3].

# LIMITATIONS

Due to the selected data acquisition method, although 400 videos were captured, the dataset only contains about 16 minutes and 40 seconds of footage (~2.5 seconds per video). However, each primitive (class) has 100 different samples, which is sufficient to feed the most common learning algorithms such as deep neural networks. Considering this limitation, the dataset was structured to facilitate an easy addition of new data.

# ETHICS STATEMENT

The author posed themselves in all videos, the videos do not show identifiable features (only hands and torso), and, as such, informed consent was not necessary.

# CRediT AUTHOR STATEMENT

**Laura Duarte:** Conceptualization, Methodology, Software, Validation, Investigation, Data Curation, Writing – Original Draft, Writing – Review & Editing, Visualization; **Pedro Neto:** Conceptualization, Writing – Review & Editing, Visualization, Supervision.

---

[2] https://github.com/Robotics-and-AI/DAVIS-data-capture-system/tree/main/arduino_code
[3] https://github.com/Robotics-and-AI/DAVIS-data-capture-system/blob/main/video.mp4

## ACKNOWLEDGEMENTS

This work was supported by Fundação para a Ciência e a Tecnologia, grant number 2021.06508.BD, UIDB/00285/2020 and LA/P/0112/2020.

## DECLARATION OF COMPETING INTERESTS

The authors declare that they have no known competing financial interests or personal relationships that could have appeared to influence the work reported in this paper.